\documentclass{article}
\usepackage{spconf,amsmath,graphicx}
\usepackage{subfigure}
\usepackage{algorithm}
\usepackage{algorithmicx}
\usepackage{algpseudocode}

\title{Intelligent Health Recommendation System for Computer Users}
\name{\normalsize Qi Guo$^1$, Zixuan Wang$^2$, Ming Li$^3$ and Hamid Aghajan$^2$}
\address{\normalsize $^1$Department of Automation, Tsinghua University, Beijing, China\\
\normalsize $^2$Department of Electrical Engineering, Stanford University, Stanford, CA, USA\\
\normalsize $^3$Shanghai Dianji University, Shanghai, China\\
}
\begin{document}
\ninept
\maketitle
\begin{abstract}
The time people spend in front of computers has been increasing steadily due to the role computers play in modern society. Individuals who sit in front of computers for an extended period of time, specifically with improper postures may incur various health issues. In this work, individuals' behaviors in front of computers are studied using web cameras. By means of non-rigid face tracking system, data are analyzed to determine the 3D head pose, blink rate and yawn frequency of computer users. When combining these visual cues, a system of intelligent personal assistants for computer users is proposed.
\end{abstract}


\section{Introduction}
\label{sec:intro}
There has been an increased interest in health status monitoring of computer users during the last decade, especially when laptops with embedded web cameras are becoming more and more popular. Everyday we can see employees, students, or iTers spend most of their time sitting in front of the computers stiffly and staring at the screen continuously. Under such circumstances, only very few of them have a planned periodical mini-breaks. Consequently, a large number of computer users are suffering from musculoskeletal Repetitive Stress Injuries (RSIs) \cite{saltzman1998computer}. In order to solve this problem, an effective mechanism should be built to remind computer users in advance. As a warning, however, people will generally feel tired after long time competitive working. Therefore, fatigue can serve as a good mark for the reminding.

This work focuses on the fatigue detection and recommendation making for computer users by means of a single web camera. To achieve the aim, non-rigid face tracking method \cite{wang2008non} was firstly applied to the real-time camera video. As a result, the position data relating to the eyes, mouth and head areas were obtained. Secondly, blink detection, yawn detection and 3D head pose analysis were performed respectively on each frame to get the three fatigue features. Finally, the fuzzy logic system was used to fuse the three features and give health recommendations to computer users hereby.

There are several ongoing research work on assistive robots~\cite{Poupart05a} to help elders, and people with special needs. Instead of building a robot, the computer can serve as the best robot for posture correction and improvements for people who suffer from repetitive stress injuries, since the injuries are rooted from using the computers. The goal of our system is to reduce the health risks by providing suggestions to improve user posture and their productivity for the short and long term. An additional advantage is the strong computation power of the computers comparing to embedded system in robots. Our work has the potential to turn the computer into a personalized care taker of human beings.

Our contributions are: 1) We propose a three layer system to prevent the user from potential health issues. 2) In the tracking layer, we improve the non-rigid face tracking algorithm by removing jitters on landmarks and reinitializing the tracking automatically. 3) In the feature layer, we propose an self-adaptive blink detection method. 4) In the recommendation layer, our inference framework using fuzzy logic to combine the expert rules and user's feedbacks give users dynamic and personalized recommendations.

The paper is organized as follows: Section 2 outlines the recent work in the vision based face tracking and assistive systems. Our system and algorithm are presented in Section 3 and Section 4 and Section 5. Experimental evaluations are shown in Section 6. Finally, we conclude the paper and discuss the future work in Section 7.

\section{Related Work}
\label{sec:related}
The proposed system may serve as an intelligent assistant for computer users. This is realized by detecting the users' fatigue status using a web camera and making health recommendations for them. The related work involves the estimations of 3D head pose, blink rate and yawn frequency respectively.

\textbf{3D head pose estimation.} Gaze direction is deeply related with a user's attention \cite{matsumoto2000algorithm}. Therefore, accurate estimation of 3D head pose helps the system to judge if a user is in correct position when using the computer and then give recommendations thereby. However, there is few generic solutions for identity-invariant head pose estimation. Detailed discussion of the inherent difficulties and evolution of this field were surveyed by Murphy \cite{murphy2009head} in 2009.

\textbf{Blink rate estimation.} In order to estimate blink rate, blink should be identified firstly. Global template matching \cite{grauman2001communication} is a common method for that. First of all, eye regions are detected. After the open status of eyes is trained, a global template is formed and can be applied to compare with real eye appearance. And the blink can be finally identified. In 2002, Morris et al. \cite{morris2002blink} proposed a real-time blink detection system, in which variance maps were used to find eye-feature points. In 2005, Chau et al. \cite{chau2005real} brought forward a blink detection system using the correlation with an online template. Yang et al. \cite{yang2012robust} modeled the shape of eyes using a pair of parameterized parabolic curves, and then fit the model globally to find potential eye regions. In addition to template matching, there are still other methods regarding blink detection, which include statistical algorithm by Pan et al. \cite{pan2007eyeblink}, optical flow based algorithm by Divjak et al. \cite{divjak2009eye}, and facial feature based algorithm by Moriyama et al. \cite{moriyama2002automatic}.

\textbf{Yawn frequency estimation.} Yawn serves as an important factor for the judgement of fatigue when combined with the 3D head pose and blink rate. Mohanty et al. \cite{mohanty2009non} proposed a non-rigid estimation algorithm for yawn detection, in which the degree of lip shape deformation was quantified. Du et al. \cite{du2011kernelized} proposed a kerneled fuzzy rough sets based yawn detection algorithm. In Omidyeganeh's work, yawn was detected based on the aspect ratio of the extracted mouth area as compared with an experimentally tuned threshold \cite{omidyeganeh2011intelligent}. Abtahi's algorithm \cite{abtahi2011driver} focused on the calculation of mouth geometric feature changes.

\section{Tracking Layer}
\label{sec:sensor}
\subsection{Hardware and Software Fundamentals}
\begin{figure}[ht]
  \begin{center}
    \includegraphics[width=.45\textwidth]{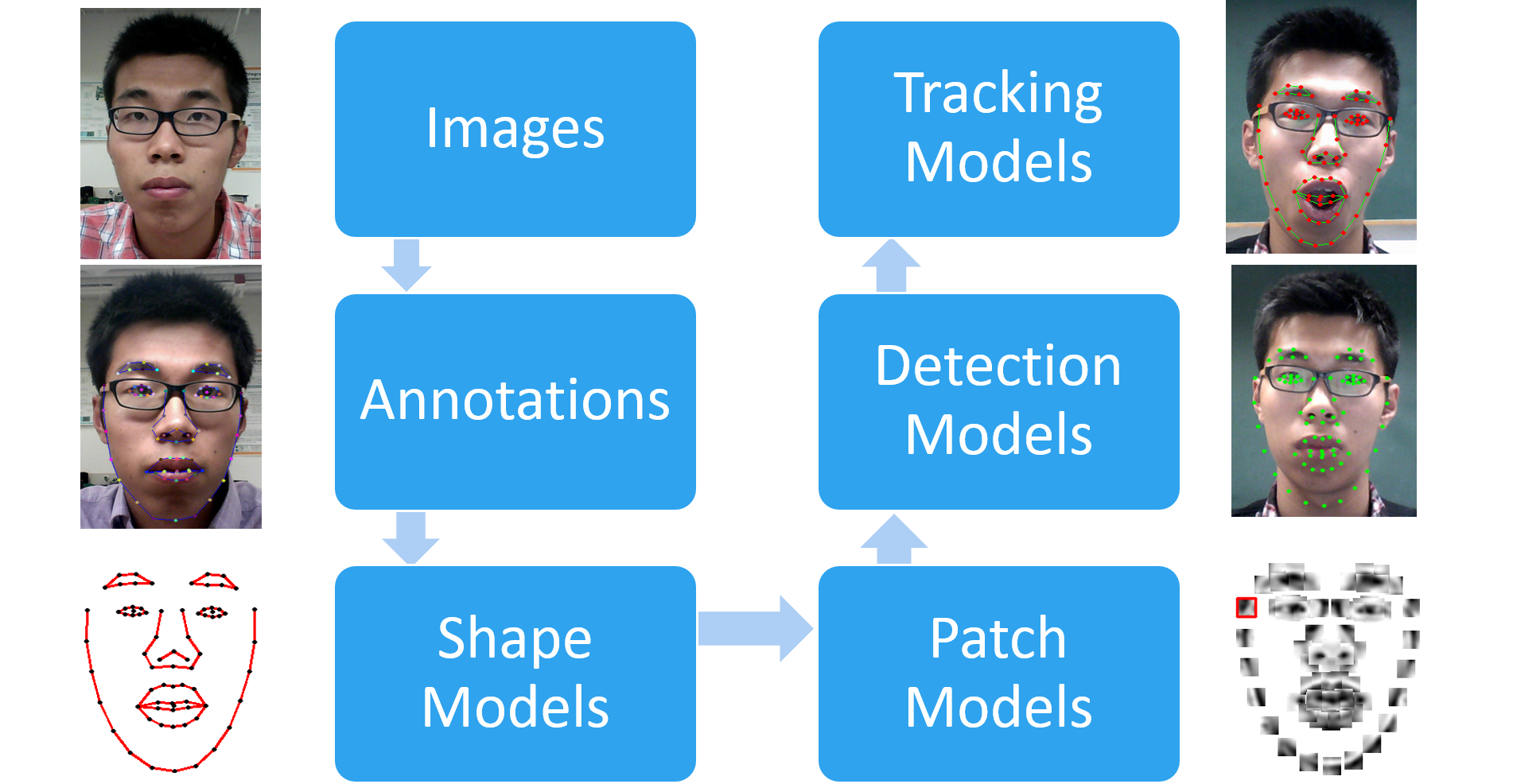}
  \end{center}
  \caption{Overview of the non-rigid face tracking system.}
  \label{fig:origin}
\end{figure}

Web camera is widely available on personal computers. It is used in our system to track facial expressional features. And a non-rigid tracking algorithm is proposed based on Active Appearance Model (AAM) \cite{cootes2001active} \cite{baggio2012mastering}. Overview of our tracking system is shown in Fig. \ref{fig:origin}. Seventy six landmarks are annotated manually on each image, from which linear shape model, correlation patch model and face detection model are trained. By combining these three model files, the tracking model is obtained and applied to track faces.

\begin{figure}[ht]
  \begin{center}
    \includegraphics[width=2.2in]{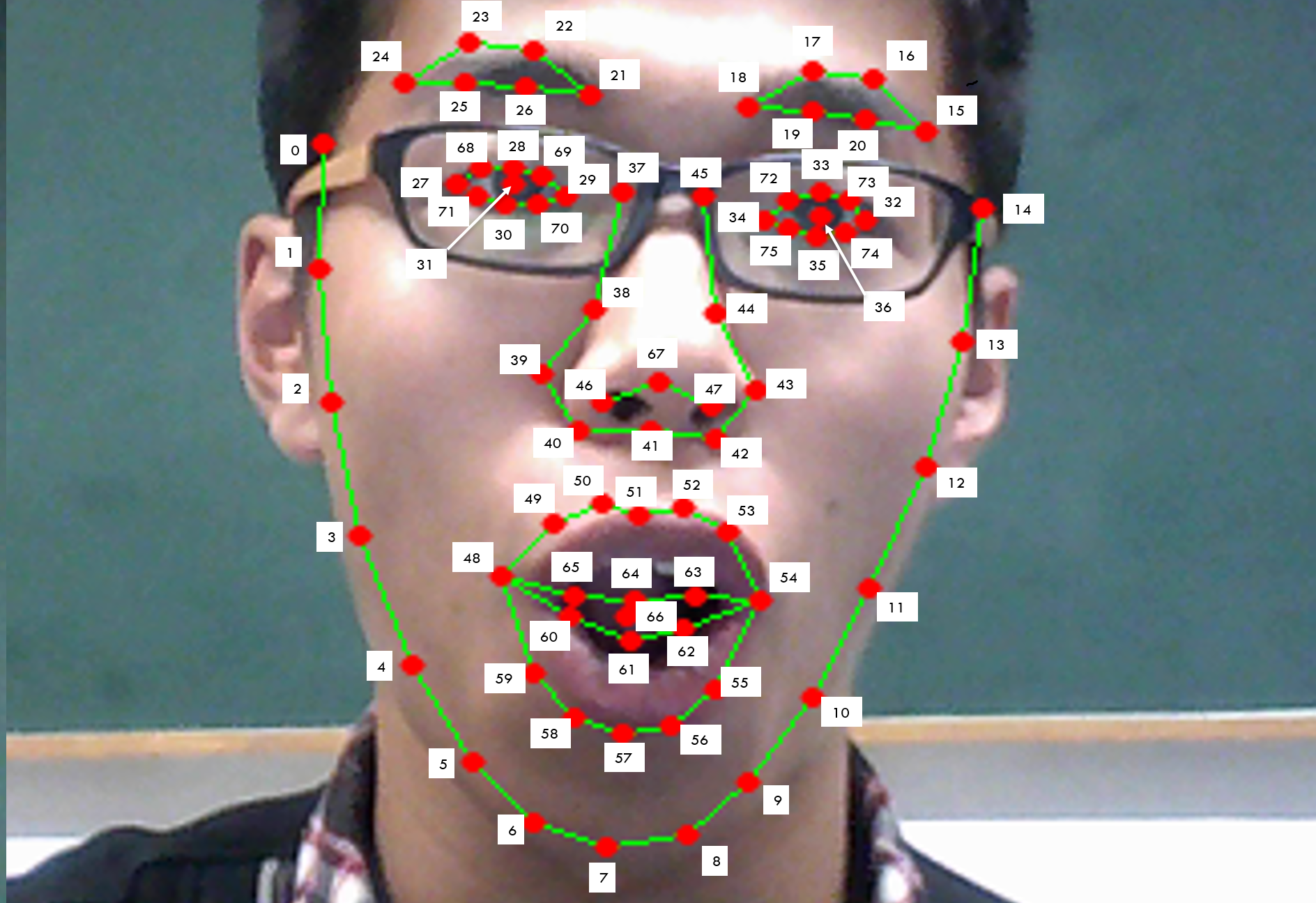}
  \end{center}
  \caption{Visualization of 76 tracking landmarks.}
  \label{fig:annotate}
\end{figure}

In order to improve the generalization performance and tracking accuracy, 385 face images with different age, expression and ethnicity groups are used for the training. Fifty eight of them are captured in our lab, and the remaining 327 images are selected out of the Biwi 3D Audiovisual Corpus of Affective Communication database \cite{fanelli20103} from the Swiss Federal Institute of Technology Zurich (ETHZ). As is shown in Fig. \ref{fig:annotate}, 76 facial landmarks are located for tracking, of which 15 lie on the chin, 6 on each eyebrow, 9 on each eye, 12 on the nose and 19 on the mouth.

\subsection{Shape and Patch Models}
Figure \ref{fig:origin} shows that shape models use a linear representation of the facial geometry to illustrate how landmarks vary across different people and expressions. The goal of linear modeling is to find a low-dimensional subspace within $2N$-dimensions ($N$ represents the total number of landmarks). Principal Component Analysis (PCA) is applied to find the best subspace. As for the correlation-based patch models, we generate a group of image patches that will produce strong responses at the exact location of landmarks based on the annotated dataset. Cross-correlation is calculated on patch models to estimate the feature locations and correct facial shape models.

The tracking procedure always suffers landmark jittering, which makes it difficult to detect facial expressional movements like blink and yawn accurately. Occlusion is another problem in the tracking. Once part of the face is occluded, reinitiating of the tracking cannot be performed automatically.
\vspace{-1em}
\subsection{Jittering Removal}
Jittering of landmarks is very common during the tracking. Therefore its pattern can be learned by means of Maximum Likelihood Estimation (MLE) and Bayesian Minimum Error Estimation (BMEE). In order to eliminate the ambiguity between facial movement and jittering, human face should keep still during the learning period for about 0.5 second in our system.

Let $\omega_i$ be the event that the $i$th landmark jitters, and $d_i$ represent the offset of the $i$th landmark in the consecutive frames ($i=1,2,...,N$), where $N$ is the total number of landmarks. Supposing that landmark jittering agrees with the Gaussian distribution, that is $p(d_i|\omega_i) \sim N(\mu_i,\sigma_i)$. We have
\begin{equation}
 \mu_i = \frac{1}{M}\sum_{k = 1}^{M}d_i^{(k)},  \sigma_i = \frac{1}{M}\sum_{k = 1}^{M}(d_i^{(k)}-\mu_i)^2,
\label{eqn:mle}
\end{equation}
where $M$ stands for the total number of the training frames.

According to Bayesian Minimum Error Estimation, instead of comparing the values of the posterior probability density $p(\omega_i|d)$ and $p(\overline{\omega_i}|d)$, we can compare the values of $p(d|\omega_i)p(\omega_i)$ and $p(d|\overline{\omega_i})p(\overline{\omega_i})$, which is equivalent to comparing the values of $p(d|\overline{\omega_i})$ with $p(d|\overline{\omega_i})p(\overline{\omega_i})/p(\omega_i)$. Given that $p(\omega_i)$ and $p(\overline{\omega_i})$ are relatively fixed, and its difficult to get $p(d|\overline{\omega_i})$, we can replace $p(d|\overline{\omega_i})p(\overline{\omega_i})/p(\omega_i)$ with a fixed threshold $\alpha$. Therefore, the classifier can be defined as the following rules:

R1: IF $p(\omega_i|d)<\alpha$, THEN the offset is caused by facial movement.

R2: IF $p(\omega_i|d)\geq \alpha$, THEN the $i$th tracking point jitters.

Let $\phi(x_1)=\alpha, \phi(x_2)=\alpha, x_1 \leq x_2$, where $\phi(x)$ and $\Phi(x)$ stand for the standard normal probability density function and cumulative distribution function respectively.
We have the error rate $P_e = \Phi(x_2)-\Phi(x_1)$. Performance of the jittering removal is evaluated in section 6.
\vspace{-1em}
\subsection{Automatic Reinitiating of Face Tracking}
\begin{figure}[ht]
\centering
\subfigure[Without Auto-reinitiating]{
 \includegraphics[width=.45\textwidth]{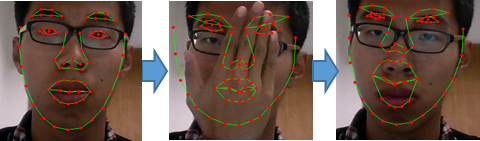}
 \label{fig:redetect:a}   
 }
 \subfigure[With Auto-reinitiating]{
 \includegraphics[width=.45\textwidth]{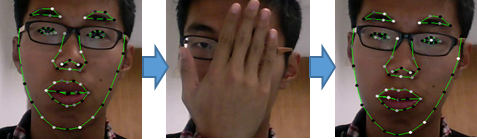}
  \label{fig:redetect:b}  
 }
  \vspace{-1em}
 \caption{Comparison of the tracking effect between with and without auto-reinitiating.}
 \label{fig:redetect}
   \vspace{-1em}
 \end{figure}

 Figure \ref{fig:redetect:a} shows that once the tracker fails in detection, tracking cannot be reinitiated automatically. This causes the landmark drift phenomenon. To solve this problem, Support Vector Machine (SVM) is used to discover the gold point for reinitiating. There are two factors that will affect tracking accuracy. One is the position of the 76 landmarks. And the other is the response value obtained from template matching. Consequently, the response value and the relative ($x$, $y$) coordinates of each tracking point compose a 228 dimensional feature vector together, in which the relative coordinates are calculated by subtracting the spatial coordinates of the tracking point from its mass center coordinates. As is shown in Fig. \ref{fig:redetect:b}, when the tracker loses the face, a cascade face detection \cite{viola2001rapid} is restarted until it tracks the objects successfully. What is more, the model training and auto-reinitiating can be performed simultaneously in real-time. Its performance can be referred to the evaluation section.


\section{Feature Layer}
\label{sec:feature}
\subsection{3D Head Pose Classification}
The pin-hole camera model is applied in our system. And 10 facial landmarks (No. 38-44, 46, 47, 67) are selected to estimate 3D head pose, because they mostly express rigid movement. Assuming the 67th landmark is the origin in 3D space, the relative coordinates of other landmarks are obtained thereby. Then the algorithm in \cite{lepetit2009epnp} is used to estimate the camera rotation and translation parameters. The corresponding relationship between the 3D space and the image plane is defined by


\begin{equation}
s
\left[
\begin{array}{ccc}
 x\\
 y\\
 1
\end{array}
\right]
=
A
\left[R|t
\right]
\left[
\begin{array}{ccc}
 X\\
 Y\\
 Z\\
 1
\end{array}
\right],
\label{eqn:solvepnp}
\end{equation}
where $s$ is a fixed coefficient. $A$ is the camera intrinsic parameter matrix, which can be computed through calibration. $[R|t]$ is the camera extrinsic parameter matrix. $R$ is the rotation matrix and $t$ is the translation vector. $(x, y)$ represents the coordinates in the image plane, and $(X, Y, Z)$ represents the coordinates in 3D space. 3D head pose is indicated by the extrinsic parameter matrix.

It has been found in our experiment that the intrinsic parameters of most web cameras are approximately the same. Therefore, the camera intrinsic parameters are set to fixed values in our system instead of repeated calibration.

An SVM classifier is applied to determine whether the user is in a correct pose while using computer. Several classes are defined as:

Pose1: The user is not looking at or not in front of the computer.

Pose2: The user is in a correct pose.

Pose3: The user is too close to the screen.

Pose4: The user is with his/her head askew to the left.

Pose5: The user is with his/her head askew to the right.

The 6 dimensional feature vector contains the rotation and translation parameters.
The classification data obtained in this layer will be used in the following section to make recommendations for the users.

\begin{figure}[ht]
  \begin{center}
    \includegraphics[width=.45\textwidth]{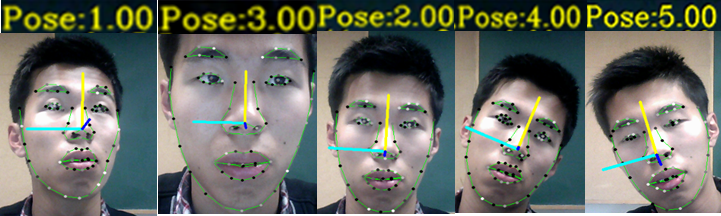}
  \end{center}
  \caption{3D head pose classification. The cyan, yellow and blue line represent the x, y and z axis respectively. The pose number 1.00-5.00 corresponds to the class C1-C5 respectively. }
  \label{fig:headposesvm}
\end{figure}

\subsection{Self-Adaptive Blink Detection}
A self-adaptive algorithm is designed to better detect the blink under different conditions, as is shown in Algorithm \ref{alg:eye}. The idea comes from the intuition that eye closure state only occupies a small proportion of the working time. And the eyeball patch color is completely different between closed and open eye state. Therefore, the average color of the eyeball patch $\overline{C}$ can be obtained and its changes can be monitored in real-time. In addition, $\overline{C}$ is normalized as $\overline{C}^\star = (\overline{C} - E\overline{C})/{\sqrt{Var\overline{C}}}$ to minimize the environmental disturbances during the tracking, where $E\overline{C}$ and $Var\overline{C}$ represent expectation and variance of $\overline{C}$. In this way, a fixed threshold $C_{t}$ can be applied to predict whether the eyes are open or closed. If the system find that the user's eye state has switched from open to closed, then blink is detected.
\begin{algorithm}\caption{$\delta = \mathrm{Eye}(I, P_l, P_r, d, w_i, h_i, C_{t}, r_f)$}
\label{alg:eye}
\begin{algorithmic}
\Require $I$ represents the image captured from the web camera. $I.r$, $I.g$ and $I.b$ represent the three color channels of the image. $P_l$ and $P_r$ represent the $(x, y)$ coordinates of the left and right eyeballs respectively, where $P_l.x$ and $P_l.y$ are the x and y coordinate respectively. $d$ denotes the distance between the face and the web camera. $w_i$ and $h_i$ are the index used to set the width and height of the patch around the eyeball. $C_{t}$ is a fixed threshold for judging the eye state. $r_f$ is the frame rate of the web camera.
\Ensure $\delta$ is the binary classification result, where 1 represents the eye open state, and 0 represents the eye closure state.\\
\State $E\overline{C} = 0$, $Var\overline{C} = 0$
\State \textbf{total frame count:} $f_c = 0$
\State \textbf{patch size:} width $w_p = w_i/d$, height $h_p = h_i/d$
\State \textbf{average color:} $\overline{C}$ = $\frac{\sum_{i, j=0}^{w_p, h_p}[I.r(i, j) + I.g(i, j) + I.b(i, j)]}{w_ph_p}$, \\$f_c = f_c + 1$, $E\overline{C} = \frac{E\overline{C} \times r_f + \overline{C}}{r_f + 1}$, $\sqrt{Var\overline{C}} = \frac{Var\overline{C} \times r_f + (\overline{C} - E\overline{C})^2}{r_f + 1}$
\State \textbf{normalization:} $\overline{C}^\star = \frac{\overline{C}-E\overline{C}}{\sqrt{Var\overline{C}}}$
\If { $f_c > r_f$}
    \If { $\overline{C}^\star < C_{t}$}
    {$\delta = 1$, eye is open}
    \Else { $\delta = 0$, eye is closed}
    \EndIf
\Else { Initialization process with no outcome}
\EndIf
\end{algorithmic}
\end{algorithm}

\subsection{Yawn Detection}
A SVM classifier is designed to determine whether the mouth is open or closed. The SVM feature vector includes the $(x, y)$ coordinates of mouth landmarks (No. 48-66). If the mouth keeps open for a preset time threshold $t_t$, then a yawn is detected.
\begin{figure}[ht]
\begin{center}
\includegraphics[width=.30\textwidth]{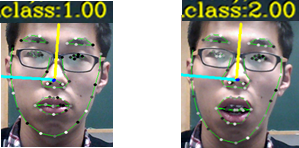}
\end{center}
\caption{Prediction of mouth condition. 1.00 represents closed mouth and 2.00 represents open mouths.}
\label{fig:mouthclass}
\end{figure}

\begin{figure*}[ht]
\centering
\includegraphics[width=\textwidth]{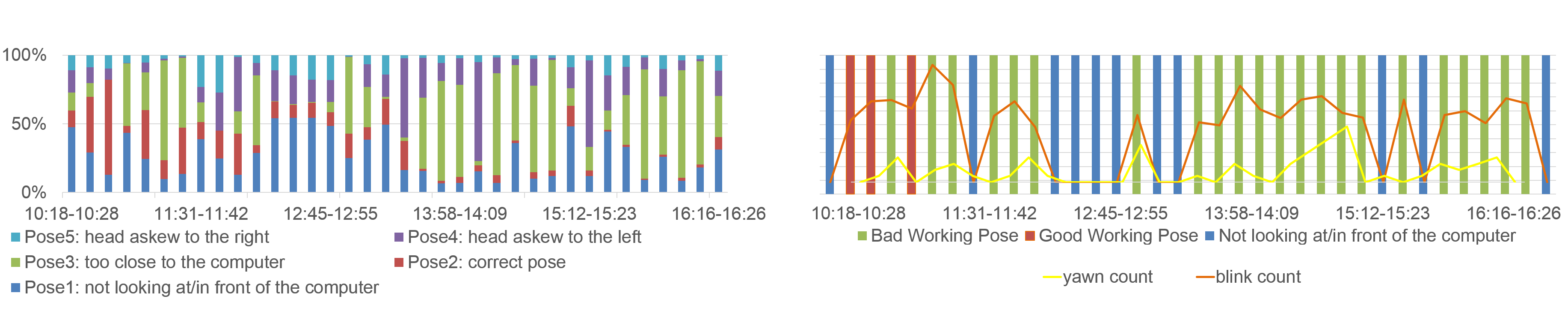}
 \caption{Outcome of a volunteer working for more than 6 hours in front of the computer. Left subfigure shows the proportion of each head pose during every ten minutes period. Right subfigure shows the prediction of the user's status, the blink and yawn counts during each period.}
 \label{fig:eva}
\end{figure*}

\section{Recommendation Layer}
\label{sec:application}
\subsection{Recommendation Framework}
After the features are gathered from the above layer, health recommendations can be made for the users in front of web cameras. Firstly, several rules are defined, which may be obtained from ergonomics experts or doctors. A few examples are given as follows:

$R_1$: IF the user works more than 30 minutes, THEN take a break.

$R_2$: IF the user keeps in a bad pose for more than 10 minutes, THEN raise the alarm.

$R_3$: IF the user yawns more than 5 times in a 10 minutes period, THEN take a break.

Each rule consists of a set of premises and a consequence, which is a recommendation generated by the system. The fuzzy logic system is used to formulate the recommendation logic in the following way:

\begin{equation}
f(x|\theta)=\frac{\sum_{i=1}^Rb_i\Pi_{j=1}^n\theta_{ij}}{\sum_{i=1}^R\Pi_{j=1}^n\theta_{ij}},
\label{eqn:base}
\end{equation}
where $R$ is the number of rules and $n$ is the number of inputs in each rule. We assume that each premise has a score, which is denoted by the confidence from the feature layer. $f(x|\theta)$ is the predicted output given an input data-tuple $x_j$. $\theta_{ij}$ is the confidence score associated with the $j$th premise in the rule $R_i$. $b_i$ is the weight associated with the rule $R_i$. All the weights need to be learned before the recommendation system works. And $f(x|\theta)$ represents the recommendation. For instance, $f(x|\theta) = 1$ means taking a break and $f(x|\theta) = -1$ means keeping working. The weights can be obtained by means of batch least square estimation.

For a rule with $n$ premises, we define:

\begin{equation}
\mu_i(x)=\Pi_{j=1}^{n}\theta_{ij},
\end{equation}

\begin{equation}
\xi_i(x)=\frac{\mu_i(x)}{\sum_{i=1}^R\mu_i(x)}.
\end{equation}
Therefore, $f(x|\theta) = b^{\mathrm{T}}\xi(x)$, where $b = (b_1, \dots, b_R)^{\mathrm{T}}$. The unknown $b$ can be easily resolved using the least squre estimation. In order to train the recommendation system, $R$ rules and at least $R$ confidence scores and corresponding actions are needed.





\subsection{Dynamic Adaptation}
Users may provide explicit or implicit feedbacks to the system. Explicit feedbacks are gathered from the users' actions as clicking a dislike button. Implicit feedbacks are learned from visual clues such as the users continue working after the system suggests to take a break. In our system, only explicit feedbacks are considered. Vector $b$ is updated by means of the Equation \ref{eqn:update}.

\begin{equation}
b^{(t)} = (1 - \alpha)b^{(t-1)} + \alpha b^{*},
\label{eqn:update}
\end{equation}
where $b^{*}$ is calculated by solving the normal equation using the users' feedback data. $\alpha$ is the adaptation rate, which means the system will gradually learn to adapt to the users' preferences.
%
%
%

\section{Evaluations}
\label{sec:evaluation}
Experiments are performed on 5 randomly picked volunteers, which includes 4 males and 1 female. They were asked to do regular computer work in front of their PCs for at least 10 minutes under different conditions. $640 \times 480$ camera resolution is tested, and the volunteers' faces may be occluded by glasses or hands. Since many people work in dim environments with computers, three of the tests are performed under the poor illuminated condition, which may lead to blur in facial features like eyes. The accuracy of blink detection, yawn detection are evaluated quantitatively and all the data are integrated to provide suggestions in our system.

As for the blink and yawn detection, we manually calculate the hitting rate and false detection rate of the system on all volunteers with the blink threshold $C_{t} = 2.5$ and mouth open time threshold $t_t = 1.5$ seconds for best performance. Ninety five blinks and twenty five yawns are taken into consideration, among which 20 blinks and 5 yawns are from each volunteer. The hitting rate are $90.5\%$ and $100\%$ for the blink and yawn respectively, and the false detection rate are $12.6\%$ and $8.3\%$ respectively.

To provide suggestions, the working time is divided into 10 minutes periods separately and the users' status in each period are determined based on the blink count, yawn count and 3D head pose. Figure \ref{fig:eva} shows the outcome of our system when tracking a volunteer working for more than 6 hours in front of the computer.
When comparing the outcome with the real condition, it can be found that the user's absence in front of the computer in 10:18-10:28, 11:31-11:42 and 13:06-13:27 periods is due to mini-breaks. The user's absence in 12:13-12:55, 15:02-15:12 and 15:23-15:33 periods is due to lunch, paper work and group discussion respectively. Moreover, the right subfigure shows that the user keeps in a bad pose most of the time and works continuously for more than 30 minutes, which will definitely trigger the system's alarm. It also shows that the user is in a potential fatigue condition during 14:30-15:00 due to the increase of yawn rate. It is because the user used to have a nap at noon but did not that day due to the experiment, which made him feel tired in the afternoon.

\section{Conclusion}
\label{sec:conclusion}
In this work, a system is presented which combines non-rigid face tracking with feature analysis to determine the working status of computer users. Jittering removal and auto-reinitiating methods are designed to improve the performances of traditional face tracking algorithms, and statistical learning methods are applied in the feature analysis. By using the blink, yawn detection and 3D head pose analysis solution, the working status of the computer users can be predicted and the recommendation rules can be made. Future work will focus on developing a user-based model to generalize the performance of the system, which will improve the tracking accuracy of tracking and feature detection.

\bibliographystyle{ieeebib}
\bibliography{refs}

\begin{thebibliography}{10}

\bibitem{saltzman1998computer}
Arthur Saltzman and CA~San~Bernardino,
\newblock ``Computer user perception of the effectiveness of exercise
  mini-breaks,''
\newblock in {\em Proceedings of the Silicon Valley Ergonomics Conference and
  Exposition}, 1998, pp. 147--151.

\bibitem{wang2008non}
Yang Wang, Simon Lucey, J~Cohn, and Jason Saragih,
\newblock ``Non-rigid face tracking with local appearance consistency
  constraint,''
\newblock in {\em IEEE international conference on automatic face and gesture
  recognition (FG¡¯08)}, 2008.

\bibitem{Poupart05a}
J.~Hoey, P.~Poupart, C.~Boutilier, and A.~Mihailidis,
\newblock ``{POMDP models for assistive technology},''
\newblock Tech. {R}ep., Proceedings of the AAAI Fall Symposium on Caring
  Machines, 2005.

\bibitem{matsumoto2000algorithm}
Yoshio Matsumoto and Alexander Zelinsky,
\newblock ``An algorithm for real-time stereo vision implementation of head
  pose and gaze direction measurement,''
\newblock in {\em Automatic Face and Gesture Recognition, 2000. Proceedings.
  Fourth IEEE International Conference on}. IEEE, 2000, pp. 499--504.

\bibitem{murphy2009head}
Erik Murphy-Chutorian and Mohan~M Trivedi,
\newblock ``Head pose estimation in computer vision: A survey,''
\newblock {\em Pattern Analysis and Machine Intelligence, IEEE Transactions
  on}, vol. 31, no. 4, pp. 607--626, 2009.

\bibitem{grauman2001communication}
Kristen Grauman, Margrit Betke, James Gips, and Gary~R Bradski,
\newblock ``Communication via eye blinks-detection and duration analysis in
  real time,''
\newblock in {\em Computer Vision and Pattern Recognition, 2001. CVPR 2001.
  Proceedings of the 2001 IEEE Computer Society Conference on}. IEEE, 2001,
  vol.~1, pp. I--1010.

\bibitem{morris2002blink}
T~Morris, Paul Blenkhorn, and Farhan Zaidi,
\newblock ``Blink detection for real-time eye tracking,''
\newblock {\em Journal of Network and Computer Applications}, vol. 25, no. 2,
  pp. 129--143, 2002.

\bibitem{chau2005real}
Michael Chau and Margrit Betke,
\newblock ``Real time eye tracking and blink detection with usb cameras,''
\newblock Tech. {R}ep., Boston University Computer Science Department, 2005.

\bibitem{yang2012robust}
Fei Yang, Xiang Yu, Junzhou Huang, Peng Yang, and Dimitris Metaxas,
\newblock ``Robust eyelid tracking for fatigue detection,''
\newblock in {\em Image Processing (ICIP), 2012 19th IEEE International
  Conference on}. IEEE, 2012, pp. 1829--1832.

\bibitem{pan2007eyeblink}
Gang Pan, Lin Sun, Zhaohui Wu, and Shihong Lao,
\newblock ``Eyeblink-based anti-spoofing in face recognition from a generic
  webcamera,''
\newblock in {\em Computer Vision, 2007. ICCV 2007. IEEE 11th International
  Conference on}. IEEE, 2007, pp. 1--8.

\bibitem{divjak2009eye}
Matjaz Divjak and Horst Bischof,
\newblock ``Eye blink based fatigue detection for prevention of computer vision
  syndrome.,''
\newblock in {\em MVA}, 2009, pp. 350--353.

\bibitem{moriyama2002automatic}
Tsuyoshi Moriyama, Takeo Kanade, Jeffrey~F Cohn, Jing Xiao, Zara Ambadar, Jiang
  Gao, and Hiroki Imamura,
\newblock ``Automatic recognition of eye blinking in spontaneously occurring
  behavior,''
\newblock in {\em Pattern Recognition, 2002. Proceedings. 16th International
  Conference on}. IEEE, 2002, vol.~4, pp. 78--81.

\bibitem{mohanty2009non}
Mihir Mohanty, Aurobinda Mishra, and Aurobinda Routray,
\newblock ``A non-rigid motion estimation algorithm for yawn detection in human
  drivers,''
\newblock {\em International Journal of Computational Vision and Robotics},
  vol. 1, no. 1, pp. 89--109, 2009.

\bibitem{du2011kernelized}
Yong Du, Qinghua Hu, Degang Chen, and Peijun Ma,
\newblock ``Kernelized fuzzy rough sets based yawn detection for driver fatigue
  monitoring,''
\newblock {\em Fundamenta Informaticae}, vol. 111, no. 1, pp. 65--79, 2011.

\bibitem{omidyeganeh2011intelligent}
Mona Omidyeganeh, Abbas Javadtalab, and Shervin Shirmohammadi,
\newblock ``Intelligent driver drowsiness detection through fusion of yawning
  and eye closure,''
\newblock in {\em Virtual Environments Human-Computer Interfaces and
  Measurement Systems (VECIMS), 2011 IEEE International Conference on}. IEEE,
  2011, pp. 1--6.

\bibitem{abtahi2011driver}
Shabnam Abtahi, Behnoosh Hariri, and Shervin Shirmohammadi,
\newblock ``Driver drowsiness monitoring based on yawning detection,''
\newblock in {\em Instrumentation and Measurement Technology Conference
  (I2MTC), 2011 IEEE}. IEEE, 2011, pp. 1--4.

\bibitem{cootes2001active}
Timothy~F. Cootes, Gareth~J. Edwards, and Christopher~J. Taylor,
\newblock ``Active appearance models,''
\newblock {\em Pattern Analysis and Machine Intelligence, IEEE Transactions
  on}, vol. 23, no. 6, pp. 681--685, 2001.

\bibitem{baggio2012mastering}
Daniel~L{\'e}lis Baggio, Shervin Emami, David~Mill{\'a}n Escriv{\'a},
  Khvedchenia Ievgen, Naureen Mahmood, Jason Saragih, and Roy Shilkrot,
\newblock {\em Mastering OpenCV with Practical Computer Vision Projects},
\newblock Packt Pub., 2012.

\bibitem{fanelli20103}
Gabriele Fanelli, Juergen Gall, Harald Romsdorfer, Thibaut Weise, and Luc
  Van~Gool,
\newblock ``A 3-d audio-visual corpus of affective communication,''
\newblock {\em Multimedia, IEEE Transactions on}, vol. 12, no. 6, pp. 591--598,
  2010.

\bibitem{viola2001rapid}
Paul Viola and Michael Jones,
\newblock ``Rapid object detection using a boosted cascade of simple
  features,''
\newblock in {\em Computer Vision and Pattern Recognition, 2001. CVPR 2001.
  Proceedings of the 2001 IEEE Computer Society Conference on}. IEEE, 2001,
  vol.~1, pp. I--511.

\bibitem{lepetit2009epnp}
Vincent Lepetit, Francesc Moreno-Noguer, and Pascal Fua,
\newblock ``Epnp: An accurate o(n) solution to the pnp problem,''
\newblock {\em International Journal of Computer Vision}, vol. 81, no. 2, pp.
  155--166, 2009.

\end{thebibliography}

\end{document}